\documentclass[conference]{IEEEtran}
\IEEEoverridecommandlockouts
\usepackage{cite}
\usepackage{amsmath,amssymb,amsfonts}
\usepackage{algorithmic}
\usepackage{graphicx}
\usepackage{textcomp}
\usepackage{xcolor}

\usepackage{algorithm}
\usepackage{algorithmic}
\usepackage{amssymb}
\usepackage{amsfonts}
\usepackage{times}
\usepackage{graphicx}
\usepackage{subcaption}
\usepackage{times,amsmath,epsfig}
\usepackage{latexsym,amssymb,mathrsfs}
\usepackage{cite}
\usepackage{color}
\usepackage{comment}
\usepackage{url}
\usepackage{epstopdf}
\usepackage{caption}

\newtheorem{assump}{Assumption}

\newtheorem{lem}{Lemma}
\newtheorem{definition}{Definition}
\newtheorem{theorem}{Theorem}

\usepackage{comment}

\def\BibTeX{{\rm B\kern-.05em{\sc i\kern-.025em b}\kern-.08em
    T\kern-.1667em\lower.7ex\hbox{E}\kern-.125emX}}
\begin{document}

\title{Robust Risk-Sensitive Reinforcement Learning \\ with Conditional Value-at-Risk}

\author{\IEEEauthorblockN{1\textsuperscript{st} Xinyi Ni}
\IEEEauthorblockA{\textit{Electrical and Computer Engineering} \\
\textit{University of California, Davis}\\
Davis, USA \\
xni@ucdavis.edu}
\and
\IEEEauthorblockN{2\textsuperscript{nd} Lifeng Lai}
\IEEEauthorblockA{\textit{Electrical and Computer Engineering} \\
\textit{University of California, Davis}\\
Davis, USA \\
lflai@ucdavis.edu}
}

\maketitle

\begin{abstract}
Robust Markov Decision Processes (RMDPs) have received significant research interest, offering an alternative to standard Markov Decision Processes (MDPs) that often assume fixed transition probabilities. RMDPs address this by optimizing for the worst-case scenarios within ambiguity sets. While earlier studies on RMDPs have largely centered on risk-neutral reinforcement learning (RL), with the goal of minimizing expected total discounted costs, in this paper, we analyze the robustness of CVaR-based risk-sensitive RL under RMDP. Firstly, we consider predetermined ambiguity sets. Based on the coherency of CVaR, we establish a connection between robustness and risk sensitivity, thus, techniques in risk-sensitive RL can be adopted to solve the proposed problem. Furthermore, motivated by the existence of decision-dependent uncertainty in real-world problems, we study problems with state-action-dependent ambiguity sets. To solve this, we define a new risk measure named NCVaR and build the equivalence of NCVaR optimization and robust CVaR optimization. We further propose value iteration algorithms and validate our approach in simulation experiments.
\end{abstract}

\begin{IEEEkeywords}
ambiguity sets, RMDP, risk-sensitive RL, CVaR
\end{IEEEkeywords}

\section{Introduction}
\label{sec:intro}
Markov Decision Processes (MDP) are foundational in Reinforcement Learning (RL), typically premised on complete knowledge of model parameters. Nevertheless, real-world applications frequently encounter uncertainties in MDP elements, such as transition probabilities and reward/cost functions, leading to estimation errors in RL algorithms and subsequent sensitivity to model inaccuracies, thus impairing performance~\cite{zhang2020robust,le2007robust,wiesemann2013robust}. In light of these challenges, Robust MDP (RMDP) has been developed to focus on optimal policies that accommodate worst-case transition probabilities within an ambiguity set~\cite{ho2018fast}, with most studies assuming known and rectangular ambiguity sets due to computational considerations~\cite{nilim2003robustness,iyengar2005robust,ben2013robust,ho2018fast,wang2021online,ho2022robust}.

\textcolor{black}{The existing RMDP research has largely focused on risk-neutral objectives that minimize the expected total discounted costs. This risk-neutral approach does not take events that are rare but have high costs into consideration. To counteract this, recently many risk-sensitive approaches where risk measures critically evaluate and quantify associated risks have been developed. Within risk-sensitive RL, the focus is on minimizing the risk of the total discounted cost to ascertain optimal policies~\cite{mihatsch2002risk}.} Coherent risk measures, conforming to principles of monotonicity, translation invariance, subadditivity, and positive homogeneity, offer a robust framework for such evaluations~\cite{artzner1999coherent}. Notably, Conditional Value-at-Risk (CVaR) has gained popularity in RL, with numerous studies proposing CVaR RL solutions for different setups~\cite{kashima2007risk,tamar2014policy,chow2014algorithms,prashanth2014policy,chow2015risk,tamar2015optimizing,chow2017risk,stanko2019risk,ma2020dsac,keramati2020being,godbout2021carl,kim2022trc,ying2022towards,lim2022distributional,greenberg2022efficient,wang2023near,zhang2024cvar}. Although risk-sensitive RL is widely popular, its robustness within the RMDP framework is not clear. While Chow et al. (2015)~\cite{chow2015risk} roughly mention how solving CVaR can enhance the robustness of risk-neutral RL in certain uncertainty sets, there is a noticeable gap in understanding how CVaR's robustness fares against various types of uncertainty sets. 

This study presents a novel and comprehensive investigation into the robustness of risk-sensitive RL within RMDP. The primary goal is to determine an optimal policy that minimizes the robust CVaR value. This value is characterized as the highest CVaR of the total discounted cost across transition probabilities within a defined rectangular ambiguity set. We initially explore scenarios where the uncertain budget is fixed, and utilize the coherent properties of CVaR and the dual representation theorem to convert the optimization challenge into a manageable risk-sensitive RL problem, facilitating the use of existing algorithms. 

Furthermore, considering that in many real-world applications, ambiguity sets are often dynamic and influenced by decision-making processes~\cite{nohadani2018optimization}, we delve deep into a more challenging setup about designing robust CVaR optimization under decision-dependent uncertainty. To tackle this problem, we introduce a new coherent risk measure NCVaR and propose a crucial decomposition theorem. We develop value iteration algorithms for NCVaR and validate our methods through simulation experiments. Based on these results, the emergence of NCVaR not only enhances the robustness of CVaR RL under decision-dependent uncertainty but also brings insights to risk-sensitive RL. Adopting NCVaR as the risk measure for risk-sensitive RL provides strong robustness compared to risk-neutral RL while rationally capturing risk. This makes NCVaR promising for potential future research and also shed lights on solving decision-dependent uncertainty for RL.

The structure of this paper is as follows: In Section~\ref{sec:pre}, we outline mathematical foundations and problem formulation. Section~\ref{sec:rbst} discusses solutions utilizing predetermined ambiguity sets and risk-sensitive RL methods. Section~\ref{sec:unfixed} focuses on undetermined ambiguity sets and corresponding value iteration algorithms. Section~\ref{sec:experiment} validates our approaches through experimental simulations and presents the numerical results. Conclusions are drawn in Section~\ref{sec:conclusion}.

\section{Preliminaries}\label{sec:pre}
\subsection{RMDP and Ambiguity Set}
We consider a MDP represented by the tuple $(\mathcal{X},\mathcal{A},C,P, \gamma,x_0)$, where $\mathcal{X}$ is the state space,  $\mathcal{A}$ denotes the action space, $C(x,a)$ specifies a bounded deterministic cost for selecting action $a$ in state $x$, $P(\cdot|x,a)$ represents the transition probability distribution, $\gamma \in [0,1]$ is the discount factor and $x_0$ denotes the given initial state. For each state $x \in \mathcal{X}$, the corresponding set of actions is represented by $\mathcal{A}(x)$. A policy $\pi$ is a mapping from the state space to the action space. The history space up to time $t \geq 1$ is represented as $H_t = H_{t-1} \times \mathcal{A} \times \mathcal{X}$, with $H_0 = \mathcal{X}$, where a history $h_t = (x_0, a_0, x_1, \dots, a_{t-1}, x_t)$ is an element of $H_t$. The policy at time $t$, $\pi_t$, maps $h_t$ to a distribution over $\mathcal{A}$. The set of such policies at time $t$ is denoted as $\Pi_{H,t}$, with $\Pi_{H} = \lim_{t\rightarrow\infty} \Pi_{H,t}$ encompassing all history-dependent policies. Similarly, $\Pi_{M,t}$ and $\Pi_{M} = \lim_{t\rightarrow\infty} \Pi_{M,t}$ denote the sets of all $t$-step and overall Markovian policies, respectively.

Addressing robustness, the transition probability $P$ is known to belong to a non-empty, compact set $\mathcal{P}$, with the uncertain transition probability denoted as $\tilde{P} \in \mathcal{P}$. The robust policy evaluation over non-rectangular ambiguity sets $\mathcal{P}$ is known to be NP-hard, even with a fixed policy $\pi$~\cite{wiesemann2013robust}. Therefore, robust RL research often focuses on rectangular ambiguity sets. In this work, we examine a specific rectangular ambiguity set:
\begin{equation*}
	\mathcal{P}=\big\{\tilde{P}: \sum_{x'\in\mathcal{X}}\tilde{P}(x'|x,a)=1, \hspace{2mm}D(\tilde{P},P)\leq K\big\},
	\label{eq:uncertain_set}
\end{equation*}
where $K$ is the non-negative uncertain budget and the divergence measure $D(\tilde{P},P)$ satisfies
\begin{equation}
	D(\tilde{P},P)=\sum_{x'\in \mathcal{X}}P(x'|x,a)\phi\big(\frac{\tilde{P}(x'|x,a)}{P(x'|x,a)}\big)\leq K.
	\label{eq:pp_dis}
\end{equation}
In~\eqref{eq:pp_dis}, $\phi:\mathbb{R}\rightarrow \mathbb{R}$ represents a convex function with the constraint $\phi(1)=0$. This function represents the $\phi$-divergence measure, a form of divergence extensively utilized in RL~\cite{gong2021f}. 
\subsection{Risk Measures}
In risk-sensitive RL, risk measures play a fundamental role in quantifying and managing risk inherent in decision-making processes. We consider a probability space $(\Omega,\mathcal{F},P)$, where $\Omega$ represents the sample space, $\mathcal{F}$ is a $\sigma$-algebra over the sample space, and $\mathbb{P}$ is a probability measure. $Z:\Omega\rightarrow\mathbb{R}$ is a bounded random variable in the probability space. 

Conditional Value-at-Risk(CVaR), which is also known as the expected shortfall or tail conditional expectation. The CVaR at confidence level $\alpha\in(0,1]$ is defined as follows~\cite{rockafellar2000optimization}:
\begin{equation*}
	\text{CVaR}_{\alpha}(Z) = \inf_{t\in \mathbb{R}}\big\{ t+\frac{1}{\alpha}  \mathbb{E}_P\left[(Z-t)^{+}\right]\big\},
	\label{eq:cvardef}
\end{equation*}
where $(z)^{+}=\max(z,0)$. One important property of CVaR is coherency and the corresponding dual representation, which serves as a crucial factor in establishing the equivalence between risk-sensitive RL and the robustness of risk-sensitive RL. The dual representation for CVaR is~\cite{ang2018dual}: $$\text{CVaR}_{\alpha}(Z)=\sup_{Q\in \mathcal{U}_{\text{CVaR}}} \mathbb{E}_{Q}[Z],$$ where $\mathcal{U}_{\text{CVaR}} = \left \{Q\ll P: D_{\text{RN}}(Q,P))\in \left[0,\frac{1}{\alpha}\right]\right \}$ with $ D_{\text{RN}}(Q,P):=\frac{Q(\omega)}{P(\omega)}$. 

We also introduce another significant risk measure, known as Entropic Value-at-Risk (EVaR). Suppose that the moment generating function $M_Z(t)=\mathbb{E}_P\left[e^{tZ}\right]$ exists for all $t\in\mathbb{R}^+$ for the random variable $Z$. In such a case, the EVaR at a given confidence level $\alpha$ is defined as follows~\cite{ahmadi2012entropic}:
\begin{equation*}
	\text{EVaR}_{\alpha}(Z)=\inf_{t>0}\left\{t^{-1}\ln(M_{Z}(t))-t^{-1}\ln\alpha\right\}. 
	\label{eq:evardef}
\end{equation*}
It is noteworthy that EVaR is also a coherent risk measure. The dual representation theorem for EVaR, as outlined in \cite{ahmadi2012entropic}, is as follows:
\begin{equation*}
	\text{EVaR}_{\alpha}(Z)=\sup_{Q\in \mathcal{U}_{\text{EVaR}}} \mathbb{E}_{Q}[Z],
	\label{eq:evardual}
\end{equation*}
where
$\mathcal{U}_{\text{EVaR}} = \left \{Q\ll P:D_{\text{KL}}(Q,P)\leq -\ln\alpha\right \}$ with $D_{\text{KL}}(Q,P):=\sum_{\omega}Q(\omega)\log\frac{Q(\omega)}{P(\omega)}$. 
\section{Robust CVaR Optimization with Predetermined Ambiguity Set}\label{sec:rbst}
In this work, the robust CVaR value is defined as the worst-case CVaR value of a policy $\pi$ when starting from the initial state $x_0$ and traversing through transition probabilities specified in the ambiguity set. The objective is to minimize this robust CVaR value across all history-dependent policies, as expressed by the following optimization problem:
\begin{equation}
	\min_{\pi\in\Pi_{H}}\max_{\tilde{P}\in\mathcal{P}}\text{CVaR}_\alpha\big[\lim_{T\rightarrow\infty}\sum_{t=0}^{T}\gamma^{t}C(x_t,a_t)\mid x_0,\pi\big].
	\label{eq:ori_prob}
\end{equation}
The sets $\Pi_{H}$ and $\mathcal{P}$ are both non-empty and compact. Additionally, the objective function is finite due to $\gamma<1$. Thus, the minimum and maximum values can be achieved, as guaranteed by the Weierstrass theorem in optimization theory~\cite{ho2022robust}. This theorem ensures that the optimization problem is well-defined and can be effectively solved to obtain the desired policy that minimizes the robust CVaR value under the given constraints. Contrasting with the robustness analysis of CVaR in~\cite{chow2015risk}, our approach evaluates the inner CVaR objective in Equation~\eqref{eq:ori_prob} across the entire set $\mathcal{P}$, instead of limiting the analysis to the true transition probabilities $P$ alone. This broader evaluation provides a more comprehensive analysis of the robustness of CVaR in diverse uncertain environments.

Recalling the coherent nature of CVaR as a risk measure and leveraging the dual representation theorem, the original optimization problem~\eqref{eq:ori_prob} can be reformulated as follows:
\begin{equation}
	\min_{\pi\in\Pi_{H}}\max_{\tilde{P}\in\mathcal{P}}\max_{Q\in\mathcal{U}_\text{CVaR}}\mathbb{E}_Q\big[\lim_{T\rightarrow\infty}\sum_{t=0}^{T}\gamma^{t}C(x_t,a_t)\mid x_0,\pi\big].
	\label{eq:dual_prob}
\end{equation}
where $\mathcal{U}_\text{CVaR}=\{Q\ll \tilde{P}: 0\leq Q(x'|x,a)/\tilde{P}(x'|x,a)\leq \frac{1}{\alpha}\}.$
Notice that the $'\sup'$ has been replaced by $'\max'$ since $\mathcal{U}_{\text{CVaR}}$ is convex and compact and the objective function is continuous in $Q$. 

We first focus on solving problem~\eqref{eq:dual_prob} with a predetermined ambiguity set, where the uncertain budget remains fixed for every state and action. Our approach involves combining two inner maximization problems by analyzing the divergence $D(Q,P)$. Under the assumption that the function $\phi$ in~\eqref{eq:pp_dis} is chosen such that $D(Q,P)$ remains bounded, i.e., $D(Q,P)\leq \tilde{K}$ (a condition satisfied by the divergence measure used in this paper), we show that problem~\eqref{eq:dual_prob} can be reformulated to:
\begin{equation}
	\min_{\pi\in\Pi_{H}}\max_{Q\in\mathcal{Q}}\mathbb{E}_Q\big[\lim_{T\rightarrow\infty}\sum_{t=0}^{T}\gamma^{t}C(x_t,a_t)\mid x_0,\pi\big],
	\label{eq:single_max}
\end{equation}
where $\mathcal{Q}=\left\{Q:D(Q,P)\leq \tilde{K}\right\}$ represents the uncertain transition problem set. 

This approach effectively addresses robust CVaR across diverse uncertainty sets by combining the set's divergence measure with the Radon-Nikodym derivative, forming a new envelope set for risk-sensitive RL. This strategy not only links the robustness of risk-sensitive RL with its intrinsic transformation but also provides a universal framework for evaluating CVaR's robustness. We further illustrate this approach by analyzing two specific $\phi$-divergence measures.

\subsection{Radon-Nikodym Derivative}
Firstly, we consider the scenario where $\phi$-divergence is Radon-Nikodym derivative, subject to a fixed uncertain budget for all states and actions: $D_{\text{RN}}(\tilde{P},P)=\frac{\tilde{P}(x'|x,a)}{P(x'|x,a)}\in[0,K]$, 
where $K\geq 0$ is a predetermined constant. 

Consequently, we obtain: $D_{\text{RN}}(Q,P)\in\left[0,\frac{K}{\alpha}\right].$
In this context, the original optimization problem~\eqref{eq:dual_prob} transforms into:
\begin{equation}
	\min_{\pi\in\Pi_{H}}\max_{Q\in\mathcal{U}_{\text{RN}}}\mathbb{E}_Q\big[\lim_{T\rightarrow\infty}\sum_{t=0}^{T}\gamma^{t}C(x_t,a_t)\mid x_0,\pi\big],
	\label{eq:rn_prob}
\end{equation}
where $\mathcal{U}_{\text{RN}}=\left\{Q\ll P: D_{\text{RN}}(Q,P)\in [0,\frac{K}{\alpha}]\right\}.$

Notice that solving problem~\eqref{eq:rn_prob} is equivalent to solving the following CVaR optimization problem with confidence level $\alpha'=\frac{\alpha}{K}$:
\begin{equation*}
	\min_{\pi\in\Pi_{H}}\text{CVaR}_{\alpha'}\big[\lim_{T\rightarrow\infty}\sum_{t=0}^{T}\gamma^{t}C(x_t,a_t)\mid x_0,\pi\big],
	\label{eq:cvar_prob}
\end{equation*}
which can be solve by employing CVaR value iteration algorithms proposed in~\cite{chow2015risk}.
\subsection{KL Divergence}
In this scenario, we consider that the uncertain transition probability $\tilde{P}$ is defined in the neighborhood of the true transition probability $P$ using the KL divergence, given by: $D_{\text{KL}}(\tilde{P},P)=\sum_{x'\in \mathcal{X}}\tilde{P}(x'|x,a)\log\big(\frac{\tilde{P}(x'|x,a)}{P(x'|x,a)}\big)\leq K,$
where $K\geq 0$ is a fixed value. Without loss of generality, we set $K=\ln\kappa$ with $\kappa\geq 1$. We can combine the two inner maximization problems into one, as the KL divergence of $Q$ and $P$ satisfies: $D_{\text{KL}}(Q,P)\leq -\ln\alpha+1/\alpha\ln\kappa= -\ln(\alpha/\kappa^{\frac{1}{\alpha}}). $
Then, the original optimization problem~\eqref{eq:dual_prob} is transformed into:
\begin{equation}
	\min_{\pi\in\Pi_{H}}\max_{Q\in\mathcal{U}_{\text{KL}}}\mathbb{E}_Q\big[\lim_{T\rightarrow\infty}\sum_{t=0}^{T}\gamma^{t}C(x_t,a_t)\mid x_0,\pi\big],
	\label{eq:kl_prob}
\end{equation}
where $\mathcal{U}_{\text{KL}}=\left\{Q\ll P: D_{\text{KL}}(Q,P)\leq-\ln\frac{\alpha}{\kappa^{\frac{1}{\alpha}}}\right\}.$

Notice that solving problem~\eqref{eq:kl_prob} is equivalent to solving the following EVaR optimization problem with confidence level $\alpha'=\alpha/\kappa^{\frac{1}{\alpha}}$:
\begin{equation*}
	\min_{\pi\in\Pi_{H}}\text{EVaR}_{\alpha'}\big[\lim_{T\rightarrow\infty}\sum_{t=0}^{T}\gamma^{t}C(x_t,a_t)\mid x_0,\pi\big].
	\label{eq:evar_prob}
\end{equation*}
The problem could be solved by existing EVaR RL works~\cite{ni2022risk}. 
\section{Robust CVaR Optimization with Decision-Dependent Uncertainty}\label{sec:unfixed}
In real-world scenarios, ambiguity sets can dynamically change due to decisions made during optimization, introducing endogenous uncertainty~\cite{luo2020distributionally}. This variability means that the uncertain budget can fluctuate over time, adding complexity to robust CVaR optimization analysis. To tackle this decision-dependent uncertainty, we focus on the Radon-Nikodym derivative, i.e.,
\begin{equation*}
	D_{\text{RN}}(\tilde{P},P)=\frac{\tilde{P}(x'|x,a)}{P(x'|x,a)}\in\left[0,\vec{\kappa}(x,a)\right],\forall (x,a) \in\mathcal{X}\times\mathcal{A},
\end{equation*}
where $\vec{\kappa}:=\left\{\vec{\kappa}(x,a),\forall s\in\mathcal{S}, a\in\mathcal{A}\right\}$ is the decision-dependent uncertainty budget vector.

By combining the dual representation theorem of CVaR, we obtain the following expression:
\begin{equation*}
	D_{\text{RN}}(Q,P)=\frac{Q(x'|x,a)}{P(x'|x,a)}\in\big[0,\frac{\vec{\kappa}(x,a)}{\alpha}\big],\forall (x,a) \in\mathcal{X}\times\mathcal{A}.
\end{equation*}
The problem at hand cannot be straightforwardly addressed by treating it as a fixed confidence level CVaR optimization. To overcome this challenge, we introduce a novel risk measure called NCVaR, which incorporates both the confidence level $\alpha$ and an undetermined uncertain budget vector $\vec{\kappa}$. Before delving into its definition, we set forth an assumption to ensure that both NCVaR and the uncertain budget are meaningful.

\begin{assump}\label{asu:budget}
	The undetermined uncertain budget satisfies $1\leq\vec{\kappa}(x,a)\leq K_{\text{max}},\forall x\in\mathcal{X}$ and $a\in\mathcal{A}$. Here $K_{\max}\geq 1$ is a real value. 
\end{assump}
We now present the formal definition of NCVaR.
\begin{definition}\label{def:NCVaR}
	For a random variable $Z:\Omega\rightarrow\mathbb{R}$ with probability mass function (p.m.f.) $P$, the NCVaR at a given confidence level $\alpha\in(0,1]$ with an undetermined uncertain budget $\vec{\kappa}$ is defined as follows:
	\begin{equation}
		\text{NCVaR}_{\alpha,\vec{\kappa}}(Z)=\sup_{Q\in\mathcal{Q}}\mathbb{E}_Q[Z],
		\label{eq:ncvar_def}
	\end{equation}
	where $\mathcal{Q}=\left\{Q: D_{\text{RN}}(Q,P)=\frac{Q(\omega)}{P(\omega)}\in\big[0,\frac{\vec{\kappa}(\omega)}{\alpha}\right],\forall \omega\in\Omega\big\}.$
\end{definition}
It's easy to observe that when $P(\omega)=0$, it implies $Q(\omega)=0$, indicating that $Q$ is absolutely continuous with respect to $P$ (i.e., $Q\ll P$). By leveraging Theorem 3.2 in~\cite{ahmadi2012entropic}, we can demonstrate that NCVaR is a coherent risk measure, which provides a solid theoretical foundation for employing NCVaR in practical applications and risk-sensitive RL scenarios. 

As a consequence of the coherency property, solving problem~\eqref{eq:single_max} with an undetermined uncertain budget defined by the Radon-Nikodym derivative is equivalently transformed into:
\begin{equation}
	\min_{\pi\in\Pi_{H}}\text{NCVaR}_{\alpha,\vec{\kappa}}\big[\lim_{T\rightarrow\infty}\sum_{t=0}^{T}\gamma^{t}C(x_t,a_t)\mid x_0,\pi\big].
	\label{eq:ncvar_prob}
\end{equation}

Given the computational challenges associated with directly computing NCVaR, as it requires knowledge of the entire distribution of the total discounted cost, we present a decomposition theorem for NCVaR, which is key to simplifying NCVaR computation and the proof is detailed in Theorem 21 of~\cite{pflug2016time}.
\begin{theorem}\label{thm:EVaRdec}
	(NCVaR Decomposition) For any $\alpha\in(0,1]$ and $\vec{\kappa}$ satisfying Assumption~\ref{asu:budget}, the $\text{NCVaR}_{\alpha,\vec{\kappa}}$ has the following decomposition
	\begin{equation*}
		\begin{split}
			\text{NCVaR}_{\alpha,\vec{\kappa}}(Z|H_t,\pi)&=\max_{\xi\in\mathcal{U}_{\text{NCVaR}}(\alpha,\vec{\kappa}(x_t,a_t),P(\cdot|x_t,a_t))}\mathbb{E}_{P}[\xi_{x_{t+1}}\\
			&\cdot \text{NCVaR}_{\alpha\xi,\vec{\kappa}}(Z|H_{t+1},\pi)|H_t,\pi],
		\end{split}		
		\label{eq:decomp}
	\end{equation*}
	where 
	$\xi(x_{t+1})=\frac{Q(x'|x,a)}{P(x'|x,a)}\geq 0$ is in the set
	\begin{equation*}
		\begin{split}
			&\mathcal{U}_{\text{NCVaR}}(\alpha,\vec{\kappa}(x_t,a_t),P(\cdot|x_t,a_t))\\
			&=\big\{\xi: \xi(x_{t+1})\in\big[0,\frac{\vec{\kappa}(x_t,a_t)}{\alpha})\big], \\
			&\hspace{10mm}\sum_{x_{t+1}\in \mathcal{X}}\xi(x_{t+1} )P(x_{t+1}|x_t,a_t)=1\big\}.
		\end{split}
	\end{equation*}
\end{theorem} 
This decomposition theorem provides a valuable insight to NCVaR computation, effectively linking the risk measure between different states, and facilitates a more tractable approach to handling the complexity of NCVaR evaluation within risk-sensitive RL under the RMDP framework. In light of the distinct confidence levels on both sides of equation~\eqref{eq:decomp}, we introduce an augmented continuous space $\mathcal{Y}=(0,1]$ to represent the domain of confidence levels.

Accordingly, the value-function $V(x,y)$ for every $(x,y)\in\mathcal{X}\times\mathcal{Y}$ is defined as:
\begin{equation*}
	\begin{split}
		&V(x,y)\\
		&=\min_{\pi \in \Pi_H}\text{NCVaR}_{y,\vec{\kappa}}\big (\lim_{T\rightarrow \infty}\sum_{t=0}^{T}\gamma^tC(x_t,a_t)|x_0=x,\pi\big ).
	\end{split}
	\label{eq:valuefun} 
\end{equation*}
The Bellman operator $\mathbf{T}:\mathcal{X} \times \mathcal{Y} \rightarrow \mathcal{X} \times \mathcal{Y}$ is defined as: 
\begin{equation*}
	\begin{split}
		\mathbf{T}[V](x,y)=&\min_{a\in \mathcal{A}}\big[C(x,a)+\gamma \max_{\xi \in \mathcal{U}_{\text{NCVaR}}(y,\vec{\kappa}(x,a),P(\cdot|x,a))}\\
		&\sum_{x'\in \mathcal{X}}\xi(x')V(x',y\xi(x'))P(x'|x,a)\big].
	\end{split}
	\label{eq:EBO}
\end{equation*}

Lemma~\ref{lem:EBOprop} introduces some important properties for the NCVaR Bellman operator.
\begin{lem}\label{lem:EBOprop}
	The Bellman operator $\mathbf{T}$ has the following properties: P1) \textit{Monotonicity}; P2) \textit{Transition invariance}; P3) \textit{Contraction}; P4) \textit{Concavity preserving}: Suppose $yV(x,y)$ is concave in $y\in\mathcal{Y},\forall x\in\mathcal{X}$. Then the maximization problem in~\eqref{eq:ncvar_prob} is concave and $y\mathbf{T}[V](x,y)$ is also concave in y.
\end{lem}
Properties P1-P3 are similar to standard dynamic programming~\cite{bertsekas2012dynamic}, and are key to design a convergent value iteration method. P4 ensures that value-iteration updates involve concave, and thus tractable, optimization problems.

Based on Lemma~\ref{lem:EBOprop}, we are able to propose the following theorem, which demonstrates the existence of a unique fixed-point solution and outline a method for deriving an optimal policy.
\begin{theorem}\label{thm:optcd}
	The unique fixed-point solution $V^*(x,y)$ of $\mathbf{T}[V](x,y)=V(x,y)$ exists and equals to the optimal value of optimization problem \eqref{eq:ncvar_prob}, i.e.,
	\begin{equation*}
		\begin{split}
			&V^*(x,y)\\
			&=\min_{\pi \in \Pi_{H}} \text{NCVaR}_{y,\vec{\kappa}}\big (\lim_{T\rightarrow \infty}\sum_{t=0}^{T}\gamma^t C(x_t,a_t)|x_0=x,\pi\big ).
		\end{split}
		\label{eq:vopti}
	\end{equation*}
\end{theorem}
Although the problem is optimized over history-dependent policies, we demonstrate that an optimal Markov policy exists, from which the optimal history-dependent policy can be derived. Considering the easier implementation of the Markov policy, we adopt the greedy policy w.r.t $V^*(x,y)$ as the optimal policy.

We introduce Algorithm~\ref{alg:VI} to effectively solve the NCVaR optimization problem. This solution is equivalent to addressing the original problem incorporating an undetermined uncertain budget defined by the Radon-Nikodym derivative.

\begin{algorithm}[h]
\caption{Value Iteration for NCVaR}
\label{alg:VI}
\begin{algorithmic}[1]
	\FOR{ $x\in\mathcal{X}$ and $y\in\mathcal{Y}$}
	\STATE arbitrarily initialize $V_0(x,y)$ 
	\ENDFOR
	\FOR{$t=0,1,2,\dots$}
	\FOR{ $x\in\mathcal{X}$ and $y\in\mathcal{Y}$}
	\STATE $\hspace{8mm}V_{t+1}(x,y)=\mathbf{T}[V_{t}](x,y)$
	\ENDFOR
	\ENDFOR
	\STATE set $V^*(x,y)=\lim_{t\rightarrow \infty}V_t(x,y)$, then  construct $\pi^*$ as the greedy policy w.r.t $V^*(x,y)$
	
\end{algorithmic}
\end{algorithm}

However, implementing Algorithm~\ref{alg:VI} directly can be challenging due to the continuous nature of the set $\mathcal{Y}$. To address this issue, we employ a sampling approach, where we select multiple points in $\mathcal{Y}$ and subsequently utilize linear interpolation to derive the value function $V$. However, to guarantee convergence, we need to satisfy the following assumption for the initial value function $V_0$.

\begin{assump}
The initial value function $V_0(x,y)$ is continuous and bounded in $y\in\mathcal{Y}$ for any $x\in\mathcal{X}$. Also, $yV_0(x,y)$ is concave in $y\in\mathcal{Y}$.
\label{asu:iniv}
\end{assump}

Let $N(x)$ denote the number of sample points, and $Y(x)={y_1,y_2,\dots,y_{N(x)}}\in [0,1]^{N(x)}$ be the corresponding confidence level set. Notably, we have $y_1 = 0$ and $y_{N(x)}=1$. To perform linear interpolation of $yV(x,y)$, we define the interpolation function $\mathcal{I}_xV$ as follows:
\begin{equation}
\begin{split}
	&\mathcal{I}_x[V](y)\\
	&=y_iV(x,y_i)+\frac{y_{i+1}V(x,y_{i+1})-y_iV(x,y_i)}{y_{i+1}-y_i}(y-y_i),
\end{split}
\end{equation}
where $y_i$ and $y_{i+1}$ are the closest points such that $y\in[y_i,y_{i+1}]$. With this, we introduce the interpolated Bellman operator for NCVaR, denoted as $\mathbf{T}_\mathcal{I}V$:
\begin{equation}
\begin{split}
	\mathbf{T}_\mathcal{I}[V](x,y)=\min_{a\in \mathcal{A}}\big [&C(x,a)+\gamma\max_{\xi \in \mathcal{U}_{\text{NCVaR}}(y,P(\cdot|x,a))}\\
	&\sum_{x'\in \mathcal{X}}\frac{\mathcal{I}_{x'}[V](y\xi(x'))}{y}P(x'|x,a)\big ].
\end{split}
\label{eq:IEBO}
\end{equation}

An essential observation regarding the interpolated Bellman operator is that it also exhibits the properties stated in Lemma~\ref{lem:EBOprop}. This can be demonstrated by employing a similar approach as used in the proof of Lemma~\ref{lem:EBOprop}. Moreover, we present a more practical and applicable version of the value iteration algorithm in Algorithm~\ref{alg:VI_lI}. This algorithm utilizes the interpolated Bellman operator and leverages linear interpolation to achieve the near-optimal value function and near-optimal policy.
\begin{algorithm}[h]
\caption{NCVaR Value Iteration with Linear Interpolation}
\label{alg:VI_lI}
\begin{algorithmic}[1]
	\STATE choose $Y(x)$, $V_0(x,y)$ satisfying Assumption~\ref{asu:iniv}
	\FOR{$t=0,1,2,\dots$}
	\FOR{$x\in\mathcal{X}$ and $y\in\mathcal{Y}$}
	\STATE $\hspace{8mm}V_{t+1}(x,y)=\mathbf{T}_\mathcal{I}[V_{t}](x,y)$
	\ENDFOR
	\ENDFOR
	\STATE set $V^*(x,y)=\lim_{t\rightarrow \infty}V_t(x,y)$, then construct $\pi^*$ as the greedy policy w.r.t $V^*(x,y)$
\end{algorithmic}
\end{algorithm}

\section{Experiment}\label{sec:experiment}
In this study, we adopt an experimental setup that aligns with previous works~\cite{chow2015risk,ni2022risk}, ensuring comparability and consistency in our results. We use a $64\times 53$ grid world RL environment with a state space representing all positions. The agent starts at $(60,50)$, aiming to reach $(60,2)$. It can move east, south, west, or north, transitioning to adjacent states with a probability of $0.95$, or to any other neighboring state with a probability of $0.05/3$. The environment has $80$ obstacles; colliding with one incurs a $40$ cost, while safe movements cost 1. The agent's goal is to find a secure and cost-effective path. For value iteration with linear interpolation, we use $21$ sample points, following the rule $y_{i+1} = \theta y_i$ for $i =1, 2,\dots,20$.

\begin{figure}[h]
\centering
\begin{subfigure}[b]{0.24\textwidth}
	\includegraphics[width=\textwidth]{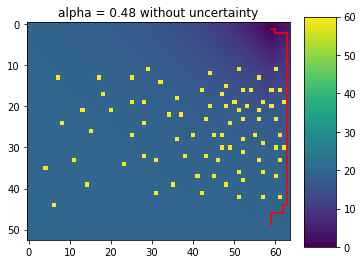}
	\caption{$\alpha=0.48$, no uncertainty}
	\label{fig:sub1}
\end{subfigure}
\begin{subfigure}[b]{0.24\textwidth}
	\includegraphics[width=\textwidth]{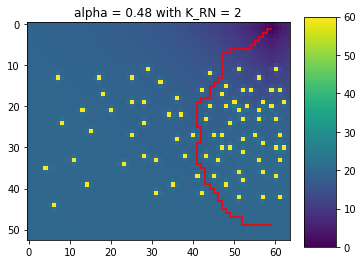}
	\caption{$\alpha=0.48$, $K_{\text{RN}}=2$}
	\label{fig:sub2}
\end{subfigure}

\begin{subfigure}[b]{0.24\textwidth}
	\includegraphics[width=\textwidth]{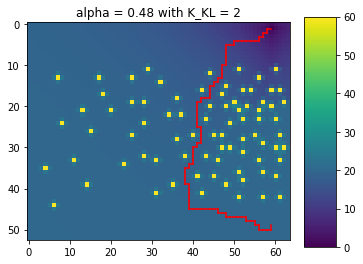}
	\caption{$\alpha=0.48$, $K_{\text{KL}}=2$}
	\label{fig:sub3}
\end{subfigure}
\begin{subfigure}[b]{0.24\textwidth}
	\includegraphics[width=\textwidth]{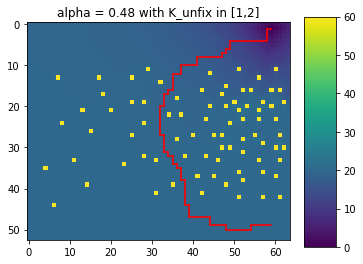}
	\caption{$\alpha=0.48$, $K_{\text{unfix}}\in[1,2]$}
	\label{fig:sub4}
\end{subfigure}

\caption{Optimal value function and path in robust CVaR optimization across various uncertainty sets.}
\label{fig:test}
\end{figure}

We first validate our approach for a fixed uncertain budget using Radon-Nikodym derivative and KL divergence. This involves visualizing the optimal value function with color variations (a bluer color indicates a lower risk while a yellower color indicates a higher risk) and tracing the optimal path as a red line (Figure~\ref{fig:sub1}). In Figure~\ref{fig:sub1},~\ref{fig:sub2} and~\ref{fig:sub3}, we select a confidence level of $\alpha=0.48$ and an uncertain budget of $K=2$ for both RN derivative and KL divergence. Consequently, we obtain $\alpha'_{\text{CVaR}}=0.24$ and $\alpha'_{\text{EVaR}}=0.03$, which indicates that the new optimal policy will exhibit a more risk-averse behavior compared to the original one. Accordingly, the optimal path becomes longer and is positioned closer to obstacles, aligning with the result that the value function is larger. We further assess Algorithm~\ref{alg:VI_lI} for decision-dependent cases, setting the uncertain budget range to $[1,2]$. As a result, for a fixed current state $x$, the new confidence level on the right side of the decomposition theorem significantly deviates from the fixed case. This increased deviation leads to the agent becoming more risk-averse as shown in Figure~\ref{fig:sub4}. In conclusion, our algorithms effectively induce risk-averse policies, equipping agents to navigate more cautiously in uncertain environments. The experiments validate our methodology's efficacy in guiding agents towards safer decision-making strategies.

\section{Conclusion and Future Direction}\label{sec:conclusion}
In this study, we have conducted a comprehensive and novel analysis of robust CVaR-based risk-sensitive RL within the framework of RMDP. We have successfully addressed robust CVaR optimization in the presence of fixed uncertain budgets while adopting a rectangular ambiguity set. We have introduced a novel risk measure NCVaR and devised NCVaR value iteration algorithms to solve the challenges associated with state-action dependent uncertainty. Furthermore, we have demonstrated the convergence of our algorithms through theoretical analysis. We have validated the proposed approaches through simulation experiments, and the results showcased the effectiveness and practicality of our methods. This paper leaves several interesting directions for future works, including the extension of robustness analysis to a broader spectrum of uncertainty sets and a deeper exploration of NCVaR within risk-sensitive RL.

\newpage
\bibliographystyle{IEEEtran}
\bibliography{IEEEfull}

\end{document}